\documentclass{article}
\usepackage{ijcai26}
\usepackage{times}
\usepackage{soul}
\usepackage[ruled,vlined,linesnumbered]{algorithm2e}
\usepackage{url}
\usepackage[hidelinks]{hyperref}
\usepackage[utf8]{inputenc}
\usepackage[small]{caption}
\usepackage{graphicx}
\usepackage{amsmath}
\usepackage{booktabs}
\usepackage[table,dvipsnames]{xcolor}
\usepackage{booktabs} 
\usepackage{amsfonts}

\usepackage{multirow} 
\usepackage[table]{xcolor}

\title{Automated Large-scale CVRP Solver Design via LLM-assisted Flexible MCTS }

\author{
    Tong Guo, ~
    Caishun Chen, ~
    Yew Soon Ong
    \affiliations
    College of Computing and Data Science,~Nanyang Technological University,~Singapore
    \emails
  tong.guo@ntu.edu.sg, ~
  chen\_caishun@a-star.edu.sg, ~
  asysong@ntu.edu.sg
}

\begin{document}

\maketitle
\renewcommand{\listalgorithmcfname}{List of Algorithms}

\begin{abstract}  
    Solving large-scale CVRP (LSCVRP) with hundreds to thousands of nodes remains difficult for even state-of-the-art solvers. 
    Divide-and-conquer can scale by decomposing the instance into size-reduced subproblems, but designing decomposition logic and configuring sub-solvers is highly expertise- and labor-intensive.
    Large Language Models (LLMs) have emerged as promising tools for automated algorithm design. However, existing LLM-driven approaches struggle with LSCVRP primarily due to the difficulty in generating sophisticated search strategies within a limited context window. 
     To bridge this gap, we propose the LLM-assisted Flexible Monte Carlo Tree Search (LaF-MCTS), a novel framework that automates the design of high-performance LSCVRP solvers. 
     We develop a three-tier decision hierarchy to enable incremental design of decomposition policies and sub-solvers for LSCVRP. 
     To enable efficient search within the algorithmic hypothesis space, we introduce \textit{semantic pruning} to eliminate semantically and structurally redundant codes, and \textit{branch regrowth} to regenerate codes and preserve diversity.
    Extensive experiments on CVRPLib demonstrate that LaF-MCTS autonomously composes and optimizes decomposition-enhanced solvers that surpasses various state-of-the-art CVRP solvers.
\end{abstract}


\section{Introduction}
As the backbone of modern supply chains, the Capacitated Vehicle Routing Problem (CVRP) is pivotal for reducing operational costs \cite{liu2023heuristics,kim2015city,zheng2025hybrid}.
With increasing real-world demands, the ability to solve large-scale CVRP (LSCVRP) with hundreds or thousands of nodes efficiently is urgent yet difficult \cite{caceres2014rich,mor2022vehicle}.
Due to its inherent NP-hard nature, the search space expands exponentially with problem size, causing even state-of-the-art CVRP solvers struggle to maintain performance on large-scale instances \cite{li2021learning,guo2025learning}.

The \textit{divide-and-conquer} strategy is an effective paradigm to tackle LSCVRP \cite{mei2011decomposition,HGS+BS,mariescu2021vrpdiv,zheng2024udc}. This strategy reduces problem dimensionality by decomposing the global problem into manageable subproblems, which are then iteratively solved by CVRP solvers. Successfully implementation relies on two core elements: a decomposition policy and a sub-solver. However, both the design of decomposition logic and the calibration of solver parameters are labor-intensive processes that rely heavily on domain expertise.

The advent of Large Language Model-assisted Algorithm Design (LLMaAD) \cite{EoH,mcts-ahd}, \cite{ReEvo}, \cite{liu2024large} has sparked a paradigm shift towards \textit{code-as-policy}, where LLMs are leveraged to automatically generate heuristic code for optimization tasks \cite{liu2024llm4ad,jiang2025large,zhang2025agentic,jiang2025droc}. This approach combines the reasoning power of LLMs with the execution speed of classical algorithms. The LLMaAD paradigm has yielded inspiring results in combinatorial optimization. Notably, FunSearch \cite{Funsearch} successfully discovered state-of-the-art heuristics for the bin packing problem, surpassing human-designed baselines. These breakthroughs provide a compelling motivation to harness the potential of LLMs for designing superior solvers specifically for LSCVRP, a domain where complexity typically defies manual optimization.

However, applying existing LLMaAD methods to LSCVRP faces two fundamental barriers. First, current methods rely on a monolithic \textit{single-thought-to-code} paradigm. While these approaches succeed in discovering simple constructive heuristics,  they become less effective for generating LSCVRP solvers, which demand a sophisticated multi-component architecture. 
Relying solely on the LLMs' internal priors and limited context windows, existing LLMaAD methods often struggle to synthesize such complex architectures directly from high-level thoughts.
Second, the prohibitive evaluation cost creates a severe bottleneck. Validating candidate solvers on large-scale instances is computationally expensive. Without an efficient search mechanism, this high cost restricts exploration within the algorithmic hypothesis space,  making it difficult to discover superior algorithms within a reasonable timeframe.

To address the limitations, we propose \textbf{LaF-MCTS} for LSCVRP.
Compared with existing LLMaAD methods, the core idea of LaF-MCTS is to mimic how human experts construct complex heuristics. 
That is, instead of forcing LLMs to design an entire solver from a single high-level thought, LaF-MCTS builds the solver progressively from global framework to fine-grained details. To achieve this idea, the innovative designs of LaF-MCTS are as follows:

\begin{enumerate}
 \item A novel three-tier decision hierarchy derived from the Decomposed Hybrid Genetic Search (HGS) algorithms \cite{HGS+BS}. 
 This hierarchy can serve as a plug-and-play algorithmic template, enabling existing LLMaAD methods to generate specialized solver components within a constrained yet expressive architectural skeleton, thereby constructing better LSCVRP solvers.

 \item  A \textit{topology-flexible} Monte Carlo Tree Search that incrementally explores the algorithmic and parameter spaces across the hierarchical tiers. 
 At each tier, LLMs are leveraged as code generators to produce diverse candidate solver components. However, LLMs often generate code snippets that are syntactically different yet semantically similar, leading to redundancy and expensive evaluation bottleneck. To solve this issue, a key design is the smeantic pruning and branch regrowth mechanism:
 \begin{itemize}
 	\item  \textit{Semantic Pruning}, which identifies and discards the semantically and structurally redundant codes generated from LLMs to reduce unnecessary evaluations.
 	\item \textit{Branch Regrowth}, which replaces pruned nodes with LLMs' regenerated logically distinct codes via negative constraint prompts to enhance exploration diversity.
 \end{itemize}

 \end{enumerate}
 
Experimental results on the well-established CVRPLib show that LaF-MCTS automatically composes and optimizes decomposition-enhanced solvers that outperform various state-of-the-art CVRP solvers, including metaheuristics, Neural Combinortorial Optimization (NCO) and LLMaAD methods.

\begin{figure*}[t!]
	\centering
	\includegraphics[width=0.95\textwidth]{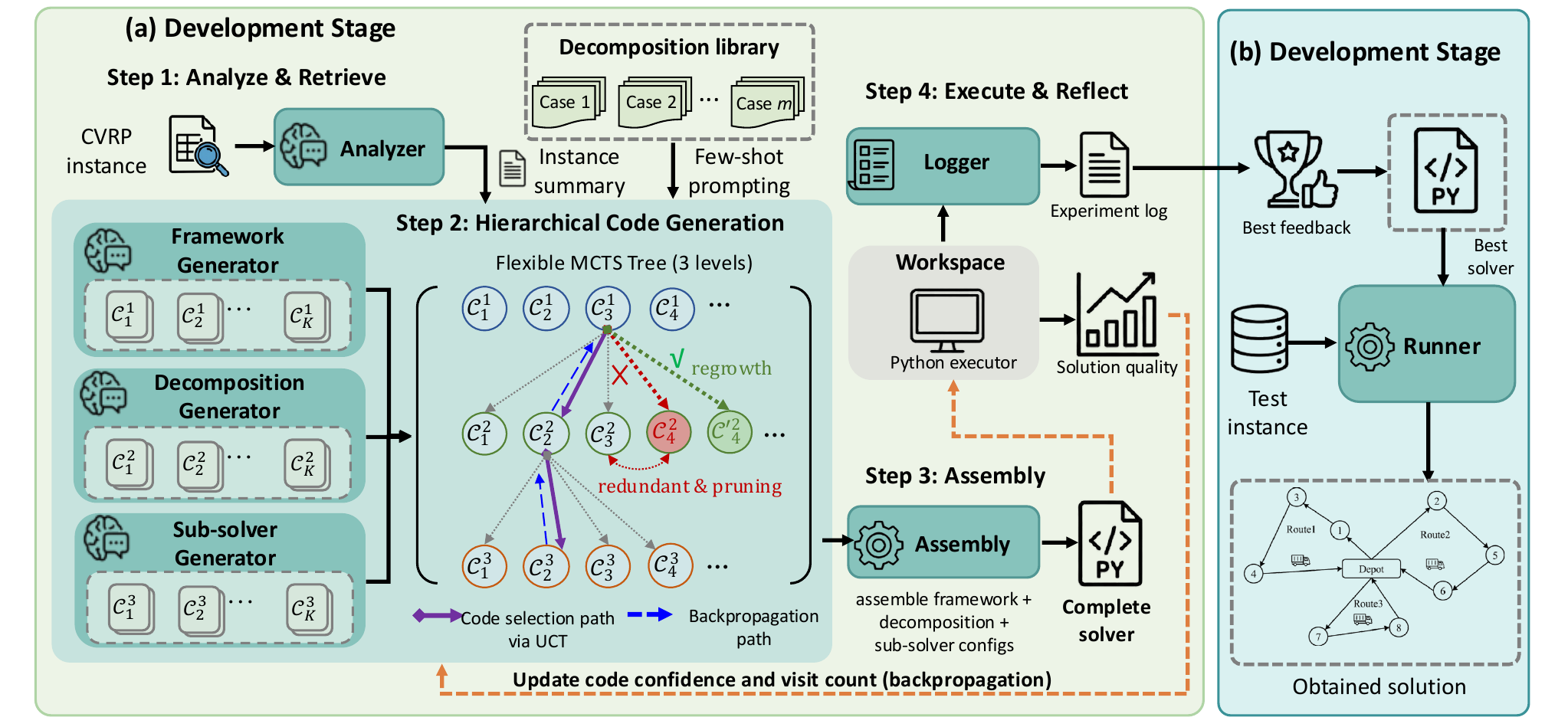} 
	\caption{The diagram of LaF-MCTS. 
    The three-tier decision hierarchy leverages LLMs to generate component codes for the framework, decomposition strategy, and sub-solver configuration. 
    These components are assembled and evaluated on training instances within a closed-loop search. 
    The code pool is dynamically evolved via iterative pruning and regrowth to ensure diversity and efficiency.
    } 
	\label{fig:framework}       
\end{figure*}

\section{Background}
The CVRP is defined on a graph $G=(V, E)$ \cite{firstvrp}, where $V=\{v_0, v_1, \dots, v_N\}$ denotes a central depot $v_0$ and $N$ customers, the objective is to minimize total travel costs. Each customer $v_i$ requires a specific demand $q_i$, and vehicles $k$ possess a capacity $Q_k$. A valid solution ensures that the aggregated demand of any route does not exceed $Q_k$ \cite{kallehauge2008formulations}. 

The HGS \cite{hgs,HGS*,pyvrp} is unequivocally recognized as the state-of-the-art algorithm for CVRP. However, its performance faces challenges as problem complexity grows. Contemporary literature defines instances with a node size exceeding 200 as LSCVRP \cite{KGLS*}. To address such scales, the prevailing consensus involves a divide-and-conquer strategy, which decomposes nodes into subgroups for subsequent optimization by HGS \cite{HGS+BS}.
Despite its effectiveness, this approach presents two major hurdles. First, designing an efficient decomposition strategy is notoriously difficult, heavily relying on domain expertise. Second, aside from the decomposition itself, HGS involves numerous hyperparameters; identifying robust configurations that perform well across varying sizes necessitates extensive, time-consuming trial-and-error. 
Consequently, the automatic design of LSCVRP solvers is of critical research significance.
We provide a more detailed description of the decomposed HGS procedure and its pseudocode in the supplementary material.

\section{Methodology}
\subsection{Overview of LaF-MCTS}
\paragraph{Major Procedures.}
As illustrated in Figure~\ref{fig:framework}, the proposed LaF-MCTS operates in two stages: {Development} and {Deployment}.
In the development stage, given CVRP training instances, an LLM \emph{Analyzer} first profiles its distributional characteristics (e.g., node spatial patterns and demand statistics) and produces a concise summary report. 
Conditioned on this report, the \emph{Hierarchical Code Generation} module decomposes solver design into a three-level hierarchy: Overall Framework Design, Decomposition Strategy Design, and Sub-solver Configuration. At each level, an LLM generates $K$ candidate code components, inducing a combinatorial space of $K^3$ possible solvers. 
Since exhaustively evaluating all candidates is infeasible, we further develop a Flexible Monte Carlo Tree Search that incrementally composes solvers from coarse to fine: it selects promising framework candidates via Upper Confidence Bound applied to Trees (UCT), expands to attach decomposition code, and finally attaches sub-solver configurations to obtain a complete executable solver. 
To reduce redundant evaluations while preserving exploration diversity, we further introduce \emph{pruning} to discard functionally similar codes at each level and \emph{regrowth} to regenerate structurally and logically distinct replacements for pruned slots. 
The selected components are then assembled and executed in Python to obtain solution-quality feedback. The resulting feedback is backpropagated through the search tree to update the confidence (value estimates) and visit counts of the selected components at all levels, enabling sustained improvement over iterations. 
In the deployment stage, LaF-MCTS directly applies the best solver discovered during development to each test CVRP instance to produce a feasible solution.
A complete pseudocode description of LaF-MCTS is provided in the supplementary material.

\paragraph{Design Rationale.}
LaF-MCTS searches over \emph{solver components} (framework, decomposition, sub-solver configurations) rather than directly over solutions, which both constrains the space to valid decomposition-based HGS solvers and enables LLMs to produce diverse, instance-aware candidates at each level. Since solver evaluation is expensive and non-differentiable, MCTS with UCT prioritizes promising partial designs under a limited budget, while \textit{pruning} removes redundant candidates and \textit{regrowth} restores diversity without wasting evaluations. This combination improves sample-efficiency and exploration quality, leading LaF-MCTS to consistently discover stronger solvers for LSCVRP cases.

\subsection{Three-tier Decision Hierarchy}
\label{subsec:decomposition}
The decision hierarchy defines the LLMaAD search space over algorithmic choices and parameter configurations.
Based on the structure of decomposed HGS \cite{HGS+BS}, the decision hierarchy is divided to three levels.

\paragraph{Tier1: Overall Framework Design.} 
This level is responsible for synthesizing the global algorithmic architecture, effectively defining the skeleton of the metaheuristic based on the main loop of HGS. 
The codes at this level should include decomposition granularity parameters (the target sub-problem size, the number of sub-problems) and the structure of algorithm execution flow (evolutionary cycle, the criteria and frequency for triggering decomposition, the selection of elite individuals to guide subproblem formation, and the mechanism for reintegrating optimized sub-solutions into the global population).
The code at this level is generated by an LLM, referred to as the \textit{Framework Generator}.

\paragraph{Tier 2: Decomposition Strategy Design.}
Conditioned on the overall algorithm framework,  the codes at this level act like clustering methods to determine the spatial logic for partitioning large-scale graphs into manageable sub-problems.
A critical challenge here is that standard LLMs, often lack exposure to specialized CVRP decomposition codes that satisfy strict domain constraints, such as sub-problem homogeneity and independence required for recursive metaheuristic solving \cite{HGS+BS}. Naively prompted models tend to generate generic clustering scripts (e.g., simple K-means) that fail to preserve the legitimacy of CVRP sub-instances or integrate seamlessly with the HGS solver.
To address this, we constructed a specialized CVRP decomposition code library implementing eleven strategies from established literature. We utilize this code library as a knowledge supplement via few-shot prompting, providing the LLMs with verified code templates. 
At this level, the decomposition strategy code is generated by a few-shot–prompted LLM using the build library, referred to as the \textit{Decomposition Generator} .

\paragraph{Tier 3: Sub-solver Configuration.}
At the lowest level, the focus shifts to local adaptation. We adopt the HGS as the algorithmic template for solving the generated sub-problems. HGS performance is affected by its configuration \cite{guo2025enhanced}. 
The \textit{Sub-solver Generator} acts as a hyper-heuristic tuner, defining codes including a parameter set tailored to the specific characteristics of the sub-problems. By dynamically calibrating this full spectrum of parameters, the agent ensures that each sub-solver is precisely tuned to the scale and topology of its designated partition.


\begin{figure*}[t!]
	\centering
	\includegraphics[width=1\textwidth]{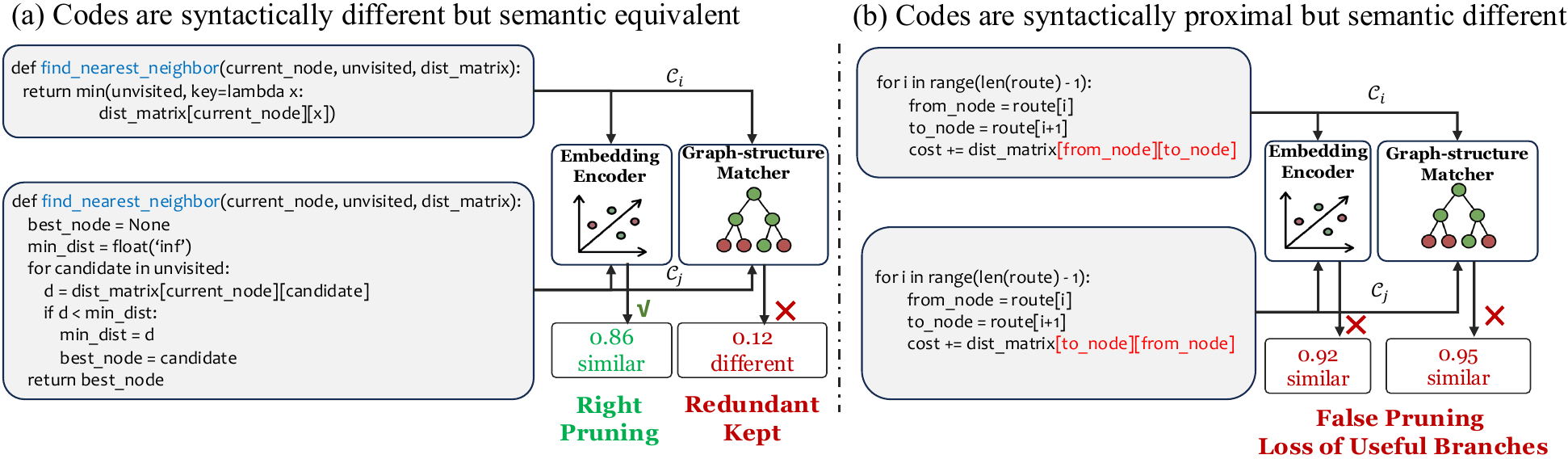} 
	\caption{The dilemma of code redundancy detection. (a) Syntactically different yet equivalent codes are well-handled by semantic embeddings but may bypass structural filters. (b) Syntactically similar but functionally divergent codes pose a risk of False Pruning, as standard metrics often fail to distinguish subtle logic changes from redundancy.}  
	\label{fig:WSMD}        
\end{figure*}

\subsection{Flexible Monte Carlo Tree Search}
LaF-MCTS constructs a search tree $\mathcal{T}$ where the root node $n_0$ represents the initial state (empty solver). The tree depth corresponds to the three-tier decision hierarchy. 
A complete path traversing from the root to the deepest leaf node represents a fully executable CVRP solver. The search process iterates through four phases: selection, selective expansion, evaluation, and backpropagation.

\paragraph{Selection.}
The search starts at the root $n_0$ and traverses down the tree to select the most promising node for expansion. At each node $n$, we employ the Upper Confidence Bound applied to Trees (UCT) criterion to balance exploration and exploitation. 
The child node $n'$ is selected as:
\begin{equation}
	n' = \underset{v \in \mathcal{C}(n)}{\arg\max} \left( \bar{\mathcal{Q}}(v) + C_p \sqrt{\frac{2 \ln \mathcal{N}(n)}{\mathcal{N}(v)}} \right)
\end{equation}
where $\mathcal{C}(n)$ is the set of child nodes of $n$, $\bar{\mathcal{Q}}(v)$ is the estimated value (average performance) of node $v$, $\mathcal{N}(n)$ and $\mathcal{N}(v)$ are the visit counts of the parent and child nodes respectively, and $C_p$ controls the exploration intensity. 

\paragraph{Selective Expansion.}
Upon reaching a leaf node $n_l$ that is not terminal, we perform expansion to generate new codes. We leverage the LLM as a generative policy $\pi_{\theta}(\cdot | s_{\text{path}})$, where $s_{\text{path}}$ is the code context accumulated along the path to $n_l$. The LLM samples $K$ distinct candidate codes $\{\mathcal{C}_1, \dots, \mathcal{C}_K\}$ for the next component.
Crucially, given the high computational cost of evaluating LSCVRP solvers, redundancy should be reduced. 
We devise a \textit{semantic pruning} and \textit{branch regrowth} mechanism to facilitate efficient tree expansion while maintaining diverse.

\paragraph{ Evaluation.}
A \textit{rollout} strategy is leveraged to estimate value of node $v$. From the newly expanded node, we perform a random walk to complete the missing downstream components until a full solver is assembled.
Then, LaF-MCTS runs the assembled solver on a fixed set of training CVRP instances. 
The reward $\mathcal{R}$ is defined as the average gap to the optimality or best-known-solutions (BKS).

\paragraph{Backpropagation.}
Once the reward $\mathcal{R}$ is computed, it is backpropagated up the tree to the root. For every node $n$ along the traversal path, the visit count and value estimates are updated:
\begin{align}
	\mathcal{N}(n) \leftarrow \mathcal{N}(n) + 1,  
	\bar{\mathcal{Q}}(n) \leftarrow \bar{\mathcal{Q}}(n) + \frac{
    \mathcal{R} - \bar{\mathcal{Q}}(n)}{\mathcal{N}(n)}
\end{align}
This iterative process gradually refines the value estimation based on the empirical performance.

\subsection{Semantic Pruning and Branch Regrowth}
LLMs often generate code snippets that are syntactically different yet semantically equivalent \cite{li2022competition}, as illustrated in Figure~\ref{fig:WSMD}(a). Evaluating such duplicate branches leads to unnecessary computational overhead. In this case, embedding-based encoders \cite{kusner2015word} can effectively identify and filter out equivalent code.
However, LLMs may also produce code snippets that are syntactically similar but semantically divergent. As shown in Figure~\ref{fig:WSMD}(b), such cases can frequently mislead both structural and embedding-based similarity metrics, resulting in the false pruning of promising branches.
To address this, we design a \textit{semantic pruning and branch regrowth} mechanism within LaF-MCTS that dynamically manages the search tree's topology.

\paragraph{Semantic Pruning via WSMD.}
To efficiently identify and prune functionally similar codes that differ in syntax, we leverage the {Word and Sentence Structure Mover's Distance} (WSMD)~\cite{yamagiwa2023improving}. Unlike superficial string matching, WSMD quantifies code similarity through a dual-process mechanism that integrates both content and logic:
(1) {Semantic embedding alignment}, which utilizes the Word Mover's Distance (WMD) to measure the semantic transport cost between token embeddings extracted from a pre-trained model (e.g., SBERT~\cite{reimers2019sentence}); and
(2) {Structural topology matching}, which employs the Gromov-Wasserstein distance to quantify discrepancies in the code's control and data flow dependencies, as captured by the Self-Attention Matrix (SAM) of the SBERT.

Formally, let $\mathbf{E}_i$ denote the sequence of token embeddings and $\mathbf{A}_i$ be the extracted SAM for a code branch $\mathcal{C}_i$. The composite distance $\mathcal{D}(\mathcal{C}_i, \mathcal{C}_j)$ is defined by the Fused Gromov-Wasserstein discrepancy:
\begin{equation}
    \mathcal{D}(\mathcal{C}_i, \mathcal{C}_j) = (1-\lambda) d_{\text{wmd}}(\mathbf{E}_i, \mathbf{E}_j) + \lambda \kappa d_{\text{smd}}(\mathbf{A}_i, \mathbf{A}_j)
\end{equation}
where $d_{\text{wmd}}(\cdot)$ penalizes semantic divergence in token usage, $d_{\text{smd}}(\cdot)$ measures the topological mismatch in dependency structures, and $\lambda$ balances the trade-off between semantic content and structural logic \cite{vayer2020fused,titouan2019optimal}. $\kappa$ is a normalization factor defined as the ratio of the average semantic cost to the average structural cost.
During the expansion phase, for any pair of generated candidate nodes $(n_i, n_j)$ representing code $\mathcal{C}_i, \mathcal{C}_j$, if $\mathcal{D}(\mathcal{C}_i, \mathcal{C}_j)  \leq \epsilon$ where $\epsilon$ is a similarity threshold, we identify them as functionally equivalent. 
Upon identifying semantically similar candidates, LaF-MCTS retains the node with the superior estimated value $\bar{\mathcal{Q}}$ and prunes the inferior variant. This helps the search budget concentrate on the most promising representative among functionally equivalent codes.

\begin{table*}[t]
	\caption{Experimental results on the CVRPLib. We report the Objective Cost (\textbf{Obj}), Optimality Gap relative to BKS (\textbf{Gap}), Best performance count across benchmarks (\textbf{BC}), Average Ranks obtained from the Friedman test (\textbf{Rank}). Best results are marked in \textbf{bold}. All methods are run with the same time budget of $N \times 2.4$ seconds as recommended by CVRPLib. Full results are provided in the supplementary material.}
	\label{tab:cvrp_results_final}
	\centering
	\footnotesize
	\setlength{\tabcolsep}{4pt} 
	\begin{tabular}{l|cc|cc|cc|cc|cc|cc}
		\toprule
		\multirow{2}{*}{\textbf{Method}} &
		\multicolumn{2}{c|}{\textbf{$N$ $\in$ [100, 200)}} &
		\multicolumn{2}{c|}{\textbf{$N$ $\in$ [200, 400)}} &
		\multicolumn{2}{c|}{\textbf{$N$ $\in$ [400, 600)}} &
		\multicolumn{2}{c|}{\textbf{$N$ $\in$ [600, 800)}} &
		\multicolumn{2}{c}{\textbf{$N$ $\in$ [800, 1000]}} & \multicolumn{2}{c}{\textbf{Statistical Metrics}} \\
		\cmidrule(lr){2-3}\cmidrule(lr){4-5}\cmidrule(lr){6-7}\cmidrule(lr){8-9}\cmidrule(lr){10-11}\cmidrule(lr){12-13}
		& Obj ($\downarrow$) & Gap ($\downarrow$) & Obj ($\downarrow$) & Gap ($\downarrow$) & Obj ($\downarrow$)& Gap ($\downarrow$) & Obj ($\downarrow$) & Gap ($\downarrow$) & Obj ($\downarrow$) & Gap ($\downarrow$) & BC ($\uparrow$) & Rank ($\downarrow$) \\
		\midrule 
		
		OR-Tools
		& 27434 & 6.60\% 
		& 50276 & 7.57\% 
		& 87469 & 6.77\% 
		& 92723 & 8.20\% 
		& 134080 & 5.30\% 
        & 0 & 5.82 \\

		HGS
		& 25754 & 0.11\% 
		& 47835 & 0.74\% 
		& 83721 & 1.21\% 
		& 87789 & 2.12\% 
		& 130449 & 2.36\%
        & 10 & 2.51 \\
		
		HGS+BS
		&  {25751} &  {0.08\%} 
		&  {47798} &  {0.67\%} 
		&  {83633} &  {1.01\%} 
		&  {87318} &  {1.54\%} 
		&  {129899} &  {1.79\%}
        &  {28} &  {2.03} \\
		\midrule
		
		MVEoE
		& 27115 & 5.2\%
		& 51365 & 8.83\%
		& 93889 & 15.44\% 
		& 100251 & 17.4\%
		& 158001 & 20.79\%
        & 0 & 8.82 \\

        RF-POMO
		& 27197 & 5.32\%
		& 50871 & 7.65\%
		& 91304 & 10.56\% 
		& 96569 & 12.31\%
		& 151332 & 15.09\%
        & 0 & 7.45 \\

         RF-TE
		& 26973 & 4.48\%
		& 50357 & 7.14\%
		& 90553 & 10.19\% 
		& 95925 & 11.62\%
		& 148777 & 13.29\%
        & 0 & 6.48 \\

        CaDA
		& 26873 & 4.14\%
		& 50426 & 7.54\%
		& 90403 & 10.89\% 
		& 96023 & 12.65\%
		& 150989 & 16.97\% 
        & 0 & 6.93 \\
        
         MoSES(RF)
		& 26936 & 4.33\%
		& 50243 & 6.78\%
		& 89973 & 9.41\% 
		& 95582 & 11.3\%
		& 147837 & 12.43\%
        & 0 & 5.96 \\

		\midrule

		FunSearch
		& 32728 & 27.55\%
		& 60538 & 28.41\%
		& 103458 & 26.29\%
		& 111066 & 30.20\%
		& 157218 & 23.87\% 
        & 0 & 12.21 \\
		
		EoH
		& 33192 & 27.42\%
		& 56234 & 21.31\%
		& 97066 & 19.44\%
		& 105584 & 21.20\%
		& 145090 & 15.65\%
        & 0 & 10.61\\
		
		ReEvo
		& 33964 & 32.62\%
		& 57177 & 25.13\%
		& 98334 & 22.13\%
		& 108998 & 27.25\%
		& 145475 & 15.44\% 
        & 0 & 10.75 \\
		
		MCTS-AHD
		& 30422 & 21.39\%
		& 54154 & 19.86\%
		& 92797 & 15.73\%
		& 100217 & 16.95\%
		& 142668 & 13.83\% 
        & 0 & 9.97 \\
        
        \midrule
		
		\textbf{LaF-MCTS (Ours)} &
         \textbf{25739} & \textbf{0.05\%}
		& \textbf{47678} & \textbf{ 0.38\% }
		& \textbf{83322} & \textbf{0.61\%}
		& \textbf{86919} & \textbf{1.10\%}
		& \textbf{129457} & \textbf{1.39\%} 
        & \textbf{74} & \textbf{1.46} \\
		
		\bottomrule
	\end{tabular}
\end{table*}

\paragraph{Branch Regrowth.}
While pruning effectively eliminates redundant search paths, it reduces the branching factor and risks insufficient exploration. To mitigate this, we design a branch regrowth mechanism applied to pruned branches. 
This process replenishes the search space by replacing discarded nodes with novel candidates.
Specifically, let $\mathcal{C}_{\text{retained}}$ denote the surviving node's code. We prompt the LLM to generate a replacement candidate, $\mathcal{C}_{\text{new}}$, using a negative constraint prompt. The prompt explicitly instructs the model: \textit{Generate a new heuristic strategy that is distinct in logic and structure from $\mathcal{C}_{\text{retained}}$.} By conditioning the generation to differ from the survivor, we force the policy $\pi_{\theta}$ to shift away from explored regions. This ensures that the LaF-MCTS tree maintains high structural diversity while escaping local optima.


\begin{figure*}[t]
	\centering
	\includegraphics[width=1\textwidth]{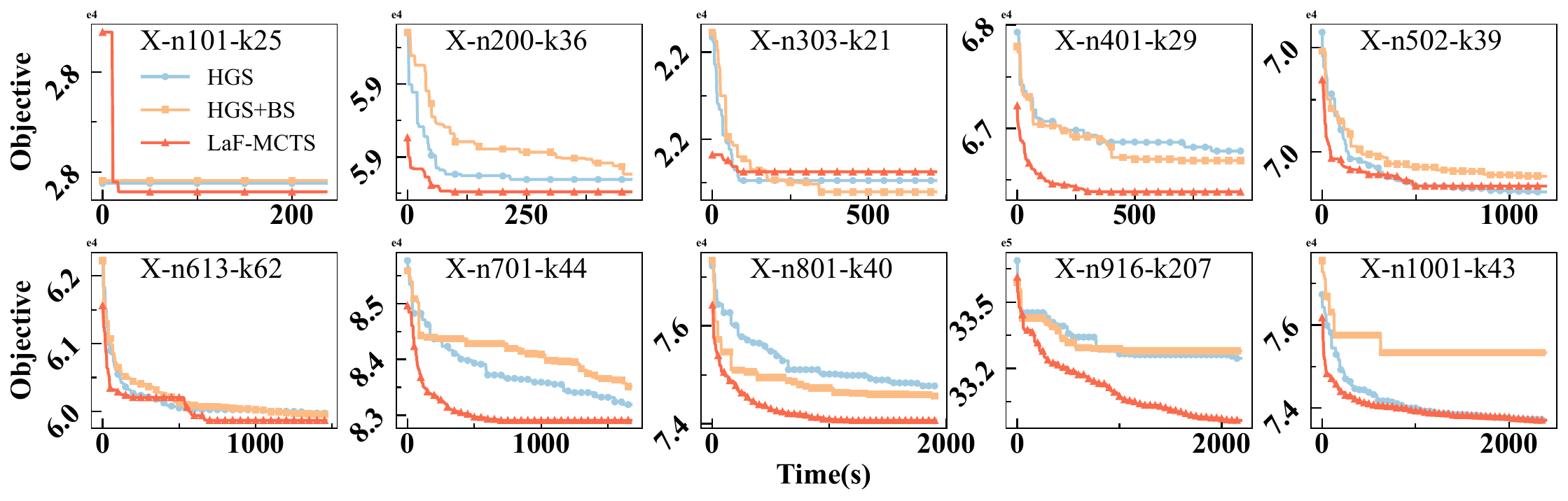} 
	\caption{Convergence curves of HGS, HGS+BS and the CVRP solver automatically designed from LaF-MCTS on representative instances.} 
	\label{fig:HGS_converge}     
\end{figure*}

\begin{table*}[t]
    \caption{Ablation experiments on the CVRPLib. We report the Objective Cost (\textbf{Obj}), Optimality Gap relative to BKS (\textbf{Gap}), Best performance count across benchmarks (\textbf{BC}), Average Ranks obtained from the Friedman test (\textbf{Rank}). Best results are marked in \textbf{bold}.} 
    \centering
    \footnotesize
    \setlength{\tabcolsep}{3pt}
    \begin{tabular}{l|cc|cc|cc|cc|cc|cc}
        \toprule
        \multirow{2}{*}{\textbf{Method}}
        &
        \multicolumn{2}{c|}{\textbf{$N$ $\in$ [100, 200)}} &
        \multicolumn{2}{c|}{\textbf{$N$ $\in$ [200, 400)}} &
        \multicolumn{2}{c|}{\textbf{$N$ $\in$ [400, 600)}} &
        \multicolumn{2}{c|}{\textbf{$N$ $\in$ [600, 800)}} &
        \multicolumn{2}{c|}{\textbf{$N$ $\in$ [800, 1000]}} &
        \multicolumn{2}{c}{\textbf{Statistical Metrics}} \\
        \cmidrule(lr){2-3}\cmidrule(lr){4-5}\cmidrule(lr){6-7}\cmidrule(lr){8-9}\cmidrule(lr){10-11}\cmidrule(lr){12-13}
        & Obj($\downarrow$) & Gap\%($\downarrow$) & Obj($\downarrow$) & Gap\%($\downarrow$) & Obj($\downarrow$) & Gap\%($\downarrow$) & Obj($\downarrow$) & Gap\%($\downarrow$)& Obj($\downarrow$) & Gap\%($\downarrow$) & BC ($\uparrow$) & Rank ($\downarrow$) \\
        \midrule
        ReEvo+T1
       & 25741 & 0.06
        & 47704 & 0.47
        & 83428 & 0.80
        & 87185 & 1.47
        & 129617 & 1.64
        & 16 & 6.74 \\

        ReEvo+T2
       & 25770 & 0.14
        & 47835 & 0.75
        & 83736 & 1.11
        & 87538 & 1.83
        & 130193 & 2.10
        & 7 & 9.30 \\

        ReEvo+Full
        & {25740} & \textbf{0.05}
        & 47702 & 0.42
        & 83389 & 0.68
        & 87114 & 1.36
        & 129872 & 1.79
        & 7 & 6.23 \\

        EoH+T1
       & \textbf{25739} & \textbf{0.05}
        & 47919 & 0.85
        & 83439 & 0.68
        & 87241 & 1.53
        & 129626 & 1.58
        & 20 & 5.60 \\

        EoH+T2
        & 25763 & 0.12
        & 47791 & 0.65
        & 83726 & 1.08
        & 87574 & 1.88
        & 129997 & 1.87
        & 6 & 8.85 \\
        
        EoH+Full
        & \textbf{25739} & \textbf{0.05}
        & 47695 & 0.42
        & {83323} & \textbf{0.61}
        & 87042 & 1.29
        & 129689 & 1.67
        & 17 & 5.30 \\ \midrule

        Fixed-Decomp
        & 25745 & 0.06
        & 47691 & 0.45
        & 83396 & 0.68
        & 87109 & 1.36
        & 129496 & 1.41
        & 15 & 5.68 \\

        Fixed-Param
          & 25756 & 0.11
        & 47779 & 0.64
        & 83731 & 1.11
        & 87663 & 1.94
        & 130150 & 2.00
        & 5 & 9.26 \\

        w/o Regrowth
       & 25779 & 0.12
        & 47815 & 0.68
        & 83702 & 1.03
        & 87184 & 1.41
        & 129937 & 1.76
        & 19 & 6.55\\

        w/o Pruning
         & 25744 & 0.07
        & 47688 & 0.39
        & 83501 & 0.85
        & 87340 & 1.70
        & 129863 & 1.83
        & 15 & 7.08  \\  

        w/o $d_{\text{wmd}}$
        & 25762 & 0.12
        & 47802 & 0.62
        & 83742 & 1.03
        & 87467 & 1.80
        & 129947 & 1.82
        & 7 &  8.33 \\

        w/o $d_{\text{smd}}$
         & 25751 & 0.08
        & 47705 & 0.47
        & 83504 & 0.79
        & 87431 & 1.76
        & 129862 & 1.83
        & 17 & 7.05 \\ \midrule


        \textbf{LaF-MCTS (Ours)}    & \textbf{25739} & \textbf{0.05}
        & \textbf{47678}     & \textbf{0.38}
        & \textbf{83322} & \textbf{0.61}
        & \textbf{86919} & \textbf{1.10}
        & \textbf{129457} & \textbf{1.39}
        & \textbf{26} & \textbf{5.05}\\
        
        \bottomrule
    \end{tabular}

    \label{tab:ablation_results}
\end{table*}

\section{Experiments}
\label{sec:experiments}

\paragraph{Datasets.}
We adopt a distinct separation between the algorithm discovery (training) phase and the evaluation phase (testing) to assess the generalization capability of the generated solvers.
For \textbf{training}, following established protocols in previous works \cite{deepaco}, we utilize synthetic datasets to drive the LLM-based search process, where the problem size is set to $500$ during the training phase.
For \textbf{testing}, we test our generated solvers on the {CVRPLib} \cite{cvrplib}. 
This benchmark is widely utilized for comparing CVRP algorithms due to its diversity in node distribution (e.g., clustered, random, quadrant) and demand types. 
The benchmark includes 100 instances with problem sizes ranging from $N=100$ to $N=1000$. Since these instances are accompanied by Best Known Solutions (BKS), they allow for an objective assessment of optimality gaps.

\paragraph{Baselines.}
We compare LaF-MCTS against three distinct categories of solvers:
(1) \textbf{Metaheuristics:} These methods represent the current best upper bound in solution quality for CVRP. We include \textbf{OR-Tools} \cite{ortools}, \textbf{HGS} \cite{pyvrp} and its decomposition-based variant \textbf{HGS-BS} (HGS with Barycenter Clustering) \cite{HGS+BS}. HGS and HGS-BS are broadly recognized as the state-of-the-art CVRP solvers.  
(2) \textbf{NCO:}  
We compare against \textbf{MVMoE} \cite{MVMoE}, \textbf{CaDA} \cite{CaDA}, \textbf{RF-POMO} \cite{liu2024multi}, \textbf{RF-TE} (modiled Transformer-based) \cite{berto2024routefinder}, and \textbf{MoSES(RF)} \cite{pan2025multi}.
RF-POMO, RF-TE, and MoSES(RF) are supported by RouteFinder \cite{berto2024routefinder}.
(3) \textbf{LLMaAD:} 
    We compare against \textbf{FunSearch} \cite{Funsearch}, \textbf{ReEvo} \cite{ReEvo},  \textbf{EoH} \cite{EoH}, and \textbf{MCTS-AHD} \cite{mcts-ahd}.

\paragraph{Settings}
Our framework utilizes \texttt{deepseek-v3.2} \cite{liu2025deepseek} as the backend LLMs. We set the number of generated branches to $K=9$ at each level and the exploration constant $C_{p}=1.0$. The semantic pruning similarity threshold is set to $\epsilon=0.5$. 
For training the LLMaAD methods, we set a same training budget of 1000 fitness evaluations. 
Preliminary experiments indicate that this budget is sufficient to ensure the convergence of the LLMaAD methods.
For testing on CVRPLib, we allocate a same time budget of $T_{max} = N \times 2.4$ seconds per instance following CVRPLib guidelines \cite{cvrplib}, ensuring full convergence of all the compared methods. 
To eliminate random bias, we perform 10 independent runs and conduct Friedman test for all experiments.

The performance metric is reported as the average objective values (\textbf{Obj}), average optimality gap (\textbf{Gap}) with respect to the BKS, best performance count across benchmarks (\textbf{BC}), average ranks obtained from the Friedman test (\textbf{Rank}).
For all the compared algorithms, we adopt the parameter values suggested in their papers, presuming that they have already been fine-tuned. 
The hardware configuration of the host consists of an Intel(R) Xeon(R) Gold 6240R CPU with 64GB of RAM and 4 GPUs of NVIDIA 3090 Ti.

\subsection{Overall Comparison}
\label{subsec:overall_comparison}

As presented in Table \ref{tab:cvrp_results_final}, our proposed LaF-MCTS demonstrates superior performance across all problem scales. In terms of solution quality, the solver discovered by our method consistently achieves the lowest average objective values and optimality gaps on all datasets ranging from $N=100$ to $N=1000$. 
The experiments indicate that, subject to the CVRPLib \cite{cvrplib} recommended time limits, metaheuristics outperform other baseline categories. 
LaF-MCTS effectively bridges the performance gap, synthesizing a solver that surpasses all baseline methods to achieve the state-of-the-art in solution quality.

Notably, HGS+BS outperforms the standard HGS, validating the necessity of decomposition strategies for handling LSCVRP complexity. 
LaF-MCTS successfully generates a decomposition-based HGS without human intervention that outperforms state-of-the-art metaheuristics in solution quality.
We plot the convergence curves of HGS and HGS+BS (the two top-performing baselines) alongside LaF-MCTS on representative instances in Figure \ref{fig:HGS_converge}.
The results show that the convergence trajectory of our method lies below that of both HGS and HGS+BS throughout the entire search process in most cases. This indicates that our generated solver outperforms HGS and HGS+BS in both effectiveness and efficiency.

Furthermore, LaF-MCTS exhibits a significant performance advantage over existing LLMaAD methods. The superior performance of LaF-MCTS over the compared LLMaAD methods stems from two primary factors: first, the proposed hierarchical design framework, coupled with specialized algorithm templates, enables the structural evolution of complex solvers, whereas typical LLMaAD methods are confined to constructive heuristics generation. 
Second, the LaF-MCTS algorithm itself demonstrates a superior capability in guiding LLMs to discover high-quality algorithms compared to existing LLMaAD paradigms, effectively outperforming both evolutionary strategies and prior tree-search approaches.
In the subsequent section, we will conduct ablation studies to verify the individual contributions of these components.

\subsection{Ablation Study}
\label{sec:ablation}

\begin{figure}[t!]
	\centering
	\includegraphics[width=0.9\linewidth]{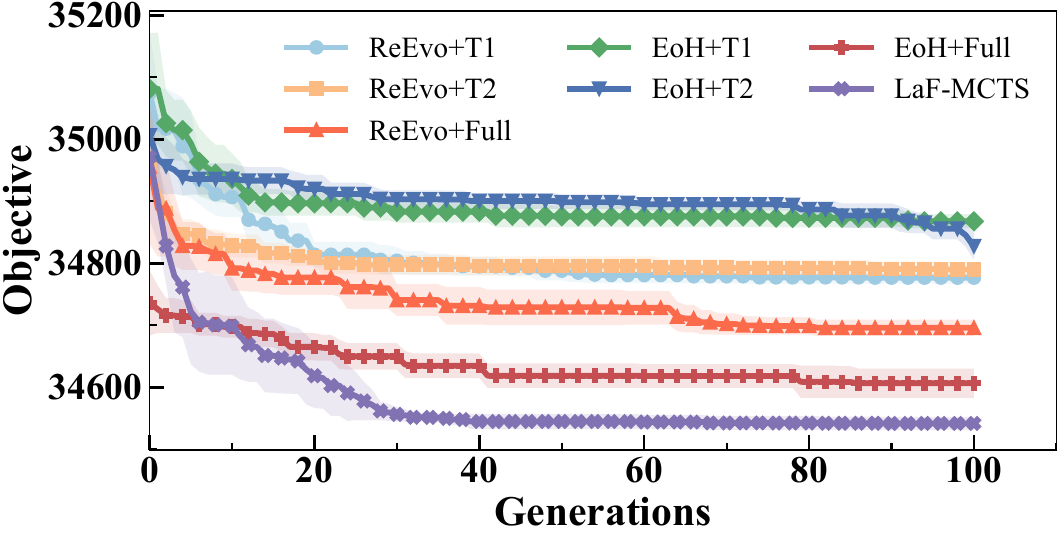} 
	\caption{Convergence curves of EoH/ReEvo variants and LaF-MCTS on the training sets. }
	\label{fig:LLMaAD_convergence}
\end{figure}

\paragraph{Effect of Three-tier Decision Hierarchy.}
We investigate if the proposed hierarchy can improve existing LLMaAD methods for generating more effective LSCVRP solvers. Using EoH/ReEvo as backbones, we compare three setups: \textbf{+T1} (Tier 1 overall framework design only), \textbf{+T2} (adds Tier 2 decomposition strategy design based on Tier 1), and \textbf{+Full} (optimizes the complete three-tier hierarchy).
Figure \ref{fig:LLMaAD_convergence} demonstrates that LaF-MCTS has better search efficiency within algorithmic hypothesis space than EoH and ReEvo variants.
Regarding generalization, LaF-MCTS has a clear lead across various problem scale. Crucially, Table \ref{tab:ablation_results} shows that fully integrated variants (\textit{+Full}) consistently outperform partial ones (\textit{+T1/T2}). 
In addition, all EoH- and ReEvo-based variants achieve substantial performance improvements over their original counterparts.
These results validate the proposed hierarchy using HGS as the algorithmic template can be utilized as a plug-and-play module to empower existing LLMaAD methods to generate superior LSCVRP solvers.

\paragraph{Effect of Parameter Tuning and Decomposition.}
Within the proposed three-tier decision hierarchy, LaF-MCTS jointly performs two key tasks: parameter tuning (for both the main framework and the sub-solver) and decomposition strategy design. To disentangle their respective contributions, we evaluate two ablated variants of LaF-MCTS:
(1) \textbf{Fixed-Decomp}, which adopts a static state-of-the-art decomposition strategy (barycenter clustering~\cite{HGS+BS}) while tuning parameters; and
(2) \textbf{Fixed-Param}, which fixes all parameters to the default HGS settings~\cite{pyvrp} while evolving decomposition strategies.
As shown in Table~\ref{tab:ablation_results}, both variants underperform the full LaF-MCTS, demonstrating the necessity of joint optimization. Moreover, {Fixed-Param} exhibits a more pronounced performance degradation than {Fixed-Decomp}, indicating that although novel decomposition strategies are beneficial, their effectiveness critically depends on adaptive parameter tuning.

\paragraph{Effect of Pruning-and-Regrowth.}
To verify if pruning-and-regrowth enhances MCTS, we evaluate: \textbf{w/o Pruning}, which removes the semantic pruning module (inherently disabling regrowth), and \textbf{w/o Regrowth}, which retains pruning but terminates redundant branches without generating diverse substitutes.
As shown in Table \ref{tab:ablation_results}, removing either module consistently degrades performance across all CVRPLib scales, highlighting their importance in guiding the LLM toward better algorithms. 
Notably, w/o Pruning performs worse than w/o Regrowth, since removing pruning also disables the regrowth trigger, reducing the framework to a standard search that lacks both redundancy filtering and adaptive diversity.

\paragraph{Effect of Semantic Pruning Metrics.}
Recall that WSMD integrates two complementary signals: semantic embedding alignment via $d_{\text{wmd}}$ and structural topology matching via $d_{\text{smd}}$. To evaluate their individual contributions, we consider two variants:
(1) \textbf{w/o $d_{\text{wmd}}$}, which removes the semantic embedding term and performs pruning solely based on structural similarity; and
(2) \textbf{w/o $d_{\text{smd}}$}, which removes the structural topology term and relies only on embedding-based similarity.
The results are summarized in Table~\ref{tab:ablation_results}. Removing either component consistently degrades performance across all benchmark instances, indicating that both semantic and structural cues are necessary for effective pruning. 
These observations empirically validate the design of WSMD, demonstrating that jointly modeling semantic content and structural logic is crucial for robust semantic pruning in LaF-MCTS.

\subsection{Further Analysis} 
\begin{figure}[t]
	\centering
	\includegraphics[width=\linewidth]{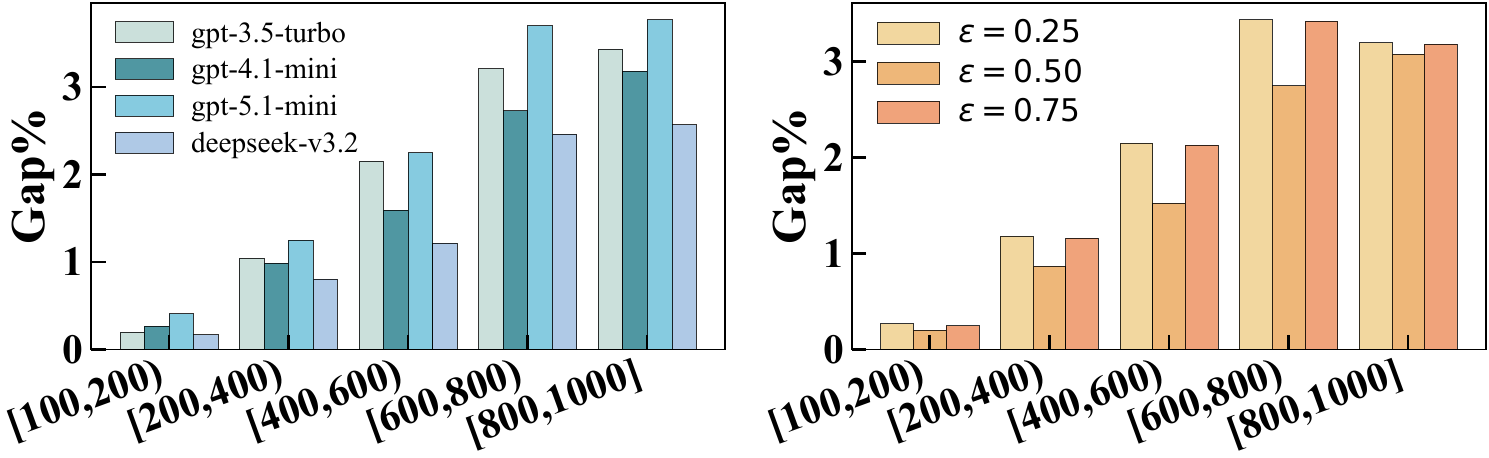}
	\caption{Parameter analysis. Left: The impact of backbone LLMs on performance. Right: The impact of $\epsilon$ on performance. }
	\label{fig:llm_analysis}
\end{figure}

We evaluate LaF-MCTS using varying backbones: \textit{GPT-3.5-Turbo}, \textit{GPT-4.1-mini}, \textit{GPT-5.1-mini}, and \textit{DeepSeek-V3.2}. As shown in Figure \ref{fig:llm_analysis} (Left), backbone choice influences performance.  \textit{DeepSeek-V3.2} achieves the lowest optimality gaps in our experiments.
The threshold $\epsilon$ in LaF-MCTS regulates the trade-off between exploration diversity and search efficiency.
Higher values risk stifling diversity by discarding distinct candidates, while lower values incur computational waste from redundancy. Figure \ref{fig:llm_analysis} (Right) shows that a conservative $\epsilon=0.50$ effectively filters redundancy while preserving structural variety, yielding the best objective values. 

In order to investigate how the effective LSCVRP solvers are built, we analyze the generated codes from LaF-MCTS at different iterative generations.
Code analysis across generations reveals that LaF-MCTS autonomously progresses from generic DBSCAN clustering to sophisticated Voronoi-based partitioning. This precise decomposition enables sub-solvers to employ expensive high-order operators (e.g., \textit{Exchange33}) that are globally intractable, demonstrating the evolution of complex strategies from simple heuristics.

\section{Conclusion}
In this paper, we proposed LaF-MCTS, a hierarchical framework that automates the design of superior solvers for large-scale Capacitated Vehicle Routing Problems (LSCVRP). By synergizing a three-tier decision hierarchy with a pruning-and-regrowth enhanced search strategy, LaF-MCTS successfully overcomes the barriers of structural complexity and evaluation costs. The discovered solvers from LaF-MCTS achieve state-of-the-art performance on CVRPLib, surpassing widely recognized baselines like HGS \cite{HGS*} and HGS+BS \cite{HGS+BS}.

LaF-MCTS currently relies on an expert-defined HGS workflow, which ensures stability but limits architectural diversity. Future work will enable fully autonomous workflow evolution. 
Furthermore, it is interesting to leverage LLMs to generate and recommend decomposition portfolios to improve generalization across data distributions.

\section{Conclusion}

\bibliographystyle{named}
\bibliography{ijcai26}

@article{HGS+BS,
  title={Decomposition strategies for vehicle routing heuristics},
  author={Santini, Alberto and Schneider, Michael and Vidal, Thibaut and Vigo, Daniele},
  journal={INFORMS Journal on Computing},
  volume={35},
  number={3},
  pages={543--559},
  year={2023},
  publisher={Informs}
}

@article{liu2023heuristics,
  title={Heuristics for vehicle routing problem: A survey and recent advances},
  author={Liu, Fei and Lu, Chengyu and Gui, Lin and Zhang, Qingfu and Tong, Xialiang and Yuan, Mingxuan},
  journal={arXiv preprint arXiv:2303.04147},
  year={2023}
}

@article{mei2011decomposition,
  title={Decomposition-based memetic algorithm for multiobjective capacitated arc routing problem},
  author={Mei, Yi and Tang, Ke and Yao, Xin},
  journal={IEEE Transactions on Evolutionary Computation},
  volume={15},
  number={2},
  pages={151--165},
  year={2011},
  publisher={IEEE}
}

@inproceedings{KGLS*,
  title={Guided local search with an adaptive neighbourhood size heuristic for large scale vehicle routing problems},
  author={Costa, Joao Guilherme Cavalcanti and Mei, Yi and Zhang, Mengjie},
  booktitle={Proceedings of the Genetic and Evolutionary Computation Conference},
  pages={213--221},
  year={2022}
}

@article{pyvrp,
  title={PyVRP: A high-performance VRP solver package},
  author={Wouda, Niels A and Lan, Leon and Kool, Wouter},
  journal={INFORMS Journal on Computing},
  volume={36},
  number={4},
  pages={943--955},
  year={2024},
  publisher={INFORMS}
}

@article{hgs,
  title={A unified solution framework for multi-attribute vehicle routing problems},
  author={Vidal, Thibaut and Crainic, Teodor Gabriel and Gendreau, Michel and Prins, Christian},
  journal={European Journal of Operational Research},
  volume={234},
  number={3},
  pages={658--673},
  year={2014},
  publisher={Elsevier}
}

@article{cvrplib,
  title={New benchmark instances for the capacitated vehicle routing problem},
  author={Uchoa, Eduardo and Pecin, Diego and Pessoa, Artur and Poggi, Marcus and Vidal, Thibaut and Subramanian, Anand},
  journal={European Journal of Operational Research},
  volume={257},
  number={3},
  pages={845--858},
  year={2017},
  publisher={Elsevier}
}

@article{kim2015city,
  title={City vehicle routing problem (city VRP): A review},
  author={Kim, Gitae and Ong, Yew-Soon and Heng, Chen Kim and Tan, Puay Siew and Zhang, Nengsheng Allan},
  journal={IEEE Transactions on Intelligent Transportation Systems},
  volume={16},
  number={4},
  pages={1654--1666},
  year={2015},
  publisher={IEEE}
}

@article{pan2025multi,
  title={Multi-task vehicle routing solver via mixture of specialized experts under state-decomposable MDP},
  author={Pan, Yuxin and Cao, Zhiguang and Gu, Chengyang and Liu, Liu and Zhao, Peilin and Chen, Yize and Lin, Fangzhen},
  journal={arXiv preprint arXiv:2510.21453},
  year={2025}
}

@inproceedings{liu2024multi,
  title={Multi-task learning for routing problem with cross-problem zero-shot generalization},
  author={Liu, Fei and Lin, Xi and Wang, Zhenkun and Zhang, Qingfu and Xialiang, Tong and Yuan, Mingxuan},
  booktitle={Proceedings of the 30th ACM SIGKDD Conference on Knowledge Discovery and Data Mining},
  pages={1898--1908},
  year={2024}
}

@article{li2022competition,
  title={Competition-level code generation with alphacode},
  author={Li, Yujia and Choi, David and Chung, Junyoung and Kushman, Nate and Schrittwieser, Julian and Leblond, R{\'e}mi and Eccles, Tom and Keeling, James and Gimeno, Felix and Dal Lago, Agustin and others},
  journal={Science},
  volume={378},
  number={6624},
  pages={1092--1097},
  year={2022},
  publisher={American Association for the Advancement of Science}
}

@misc{CaDA,
      title={CaDA: Cross-Problem Routing Solver with Constraint-Aware Dual-Attention}, 
      author={Han Li and Fei Liu and Zhi Zheng and Yu Zhang and Zhenkun Wang},
      year={2024},
      eprint={2412.00346},
      archivePrefix={arXiv},
      primaryClass={cs.AI},
      url={https://arxiv.org/abs/2412.00346}, 
}

@article{guo2025enhanced,
  title={Enhanced evolution of parallel algorithm portfolio for vehicle routing problem via transfer optimization},
  author={Guo, Tong and Mei, Yi and Zhang, Mengjie and Tang, Ke and Cai, Kaiquan and Du, Wenbo},
  journal={IEEE Transactions on Evolutionary Computation},
  year={2025},
  publisher={IEEE}
}

@article{zheng2025hybrid,
  title={Hybrid memetic search for electric vehicle routing with time windows, simultaneous pickup-delivery, and partial recharges},
  author={Zheng, Zubin and Liu, Shengcai and Ong, Yew-Soon},
  journal={IEEE Transactions on Emerging Topics in Computational Intelligence},
  year={2025},
  publisher={IEEE}
}

@article{zheng2024udc,
  title={UDC: A unified neural divide-and-conquer framework for large-scale combinatorial optimization problems},
  author={Zheng, Zhi and Zhou, Changliang and Xialiang, Tong and Yuan, Mingxuan and Wang, Zhenkun},
  journal={Advances in Neural Information Processing Systems},
  volume={37},
  pages={6081--6125},
  year={2024}
}

@article{mariescu2021vrpdiv,
  title={VRPDiv: a divide and conquer framework for large vehicle routing problems},
  author={Mariescu-Istodor, Radu and Cristian, Alexandru and Negrea, Mihai and Cao, Peiwei},
  journal={ACM Transactions on Spatial Algorithms and Systems (TSAS)},
  volume={7},
  number={4},
  pages={1--41},
  year={2021},
  publisher={ACM New York, NY}
}

@article{li2021learning,
  title={Learning to delegate for large-scale vehicle routing},
  author={Li, Sirui and Yan, Zhongxia and Wu, Cathy},
  journal={Advances in Neural Information Processing Systems},
  volume={34},
  pages={26198--26211},
  year={2021}
}

@inproceedings{jiang2025droc,
  title={DRoC: Elevating large language models for complex vehicle routing via decomposed retrieval of constraints},
  author={Jiang, Xia and Wu, Yaoxin and Zhang, Chenhao and Zhang, Yingqian},
  booktitle={13th international Conference on Learning Representations, ICLR 2025},
  year={2025},
  organization={OpenReview. net}
}

@article{zhang2025agentic,
  title={An Agentic Framework with LLMs for Solving Complex Vehicle Routing Problems},
  author={Zhang, Ni and Cao, Zhiguang and Zhou, Jianan and Zhang, Cong and Ong, Yew-Soon},
  journal={arXiv preprint arXiv:2510.16701},
  year={2025}
}

@article{jiang2025large,
  title={Large Language Models as End-to-end Combinatorial Optimization Solvers},
  author={Jiang, Xia and Wu, Yaoxin and Li, Minshuo and Cao, Zhiguang and Zhang, Yingqian},
  journal={arXiv preprint arXiv:2509.16865},
  year={2025}
}

@article{caceres2014rich,
  title={Rich vehicle routing problem: Survey},
  author={Caceres-Cruz, Jose and Arias, Pol and Guimarans, Daniel and Riera, Daniel and Juan, Angel A},
  journal={ACM Computing Surveys (CSUR)},
  volume={47},
  number={2},
  pages={1--28},
  year={2014},
  publisher={ACM New York, NY, USA}
}

@article{mor2022vehicle,
  title={Vehicle routing problems over time: a survey},
  author={Mor, Andrea and Speranza, Maria Grazia},
  journal={Annals of Operations Research},
  volume={314},
  number={1},
  pages={255--275},
  year={2022},
  publisher={Springer}
}

@article{HGS*,
  title={Hybrid genetic search for the CVRP: Open-source implementation and SWAP* neighborhood},
  author={Vidal, Thibaut},
  journal={Computers \& Operations Research},
  volume={140},
  pages={105643},
  year={2022},
  publisher={Elsevier}
}

@article{liu2025deepseek,
  title={Deepseek-v3. 2: Pushing the frontier of open large language models},
  author={Liu, Aixin and Mei, Aoxue and Lin, Bangcai and Xue, Bing and Wang, Bingxuan and Xu, Bingzheng and Wu, Bochao and Zhang, Bowei and Lin, Chaofan and Dong, Chen and others},
  journal={arXiv preprint arXiv:2512.02556},
  year={2025}
}

@inproceedings{MVMoE,
  title={MVMoE: Multi-Task Vehicle Routing Solver with Mixture-of-Experts},
  author={Zhou, Jianan and Cao, Zhiguang and Wu, Yaoxin and Song, Wen and Ma, Yining and Zhang, Jie and Xu, Chi},
  booktitle={41st International Conference on Machine Learning, ICML 2024},
  pages={61804--61824},
  year={2024},
  organization={PMLR}
}

@article{deepaco,
  title={DeepACO: Neural-enhanced ant systems for combinatorial optimization},
  author={Ye, Haoran and Wang, Jiarui and Cao, Zhiguang and Liang, Helan and Li, Yong},
  journal={Advances in neural information processing systems},
  volume={36},
  pages={43706--43728},
  year={2023}
}

@inproceedings{kusner2015word,
  title={From word embeddings to document distances},
  author={Kusner, Matt and Sun, Yu and Kolkin, Nicholas and Weinberger, Kilian},
  booktitle={International conference on machine learning},
  pages={957--966},
  year={2015},
  organization={PMLR}
}

@inproceedings{liu2024large,
  title={Large language models as evolutionary optimizers},
  author={Liu, Shengcai and Chen, Caishun and Qu, Xinghua and Tang, Ke and Ong, Yew-Soon},
  booktitle={2024 IEEE Congress on Evolutionary Computation (CEC)},
  pages={1--8},
  year={2024},
  organization={IEEE}
}

@article{vayer2020fused,
  title={Fused Gromov-Wasserstein distance for structured objects},
  author={Vayer, Titouan and Chapel, Laetitia and Flamary, R{\'e}mi and Tavenard, Romain and Courty, Nicolas},
  journal={Algorithms},
  volume={13},
  number={9},
  pages={212},
  year={2020},
  publisher={MDPI}
}

@inproceedings{titouan2019optimal,
  title={Optimal transport for structured data with application on graphs},
  author={Titouan, Vayer and Courty, Nicolas and Tavenard, Romain and Flamary, R{\'e}mi},
  booktitle={International Conference on Machine Learning},
  pages={6275--6284},
  year={2019},
  organization={PMLR}
}

@article{kallehauge2008formulations,
  title={Formulations and exact algorithms for the vehicle routing problem with time windows},
  author={Kallehauge, Brian},
  journal={Computers \& Operations Research},
  volume={35},
  number={7},
  pages={2307--2330},
  year={2008},
  publisher={Elsevier}
}

@article{guo2025learning,
  title={Learning-aided Neighborhood Search for Vehicle Routing Problems},
  author={Guo, Tong and Mei, Yi and Zhang, Mengjie and Zhao, Haoran and Cai, Kaiquan and Du, Wenbo},
  journal={IEEE Transactions on Pattern Analysis and Machine Intelligence},
  year={2025},
  publisher={IEEE}
}

@software{ortools,
  title = {OR-Tools},
  version = { v9.10 },
  author = {Laurent Perron and Vincent Furnon},
  organization = {Google},
  url = {https://developers.google.com/optimization/},
  date = { 2024-05-07 }
}

@article{liu2024llm4ad,
  title={Llm4ad: A platform for algorithm design with large language model},
  author={Liu, Fei and Zhang, Rui and Xie, Zhuoliang and Sun, Rui and Li, Kai and Lin, Xi and Wang, Zhenkun and Lu, Zhichao and Zhang, Qingfu},
  journal={arXiv preprint arXiv:2412.17287},
  year={2024}
}

@inproceedings{berto2024routefinder,
  title={RouteFinder: Towards Foundation Models for Vehicle Routing Problems},
  author={Berto, Federico and Hua, Chuanbo and Zepeda, Nayeli Gast and Hottung, Andr{\'e} and Wouda, Niels and Lan, Leon and Tierney, Kevin and Park, Jinkyoo},
  booktitle={ICML 2024 Workshop on Foundation Models in the Wild}
}

@article{mcts-ahd,
  title={Monte carlo tree search for comprehensive exploration in llm-based automatic heuristic design},
  author={Zheng, Zhi and Xie, Zhuoliang and Wang, Zhenkun and Hooi, Bryan},
  journal={arXiv preprint arXiv:2501.08603},
  year={2025}
}

@article{reevo,
  title={Reevo: Large language models as hyper-heuristics with reflective evolution},
  author={Ye, Haoran and Wang, Jiarui and Cao, Zhiguang and Berto, Federico and Hua, Chuanbo and Kim, Haeyeon and Park, Jinkyoo and Song, Guojie},
  journal={Advances in neural information processing systems},
  volume={37},
  pages={43571--43608},
  year={2024}
}

@inproceedings{yamagiwa2023improving,
  title={Improving word mover’s distance by leveraging self-attention matrix},
  author={Yamagiwa, Hiroaki and Yokoi, Sho and Shimodaira, Hidetoshi},
  booktitle={Findings of the Association for Computational Linguistics: EMNLP 2023},
  pages={11160--11183},
  year={2023}
}

@article{firstvrp,
  title={The truck dispatching problem},
  author={Dantzig, George B and Ramser, John H},
  journal={Manage. Sci.},
  volume={6},
  number={1},
  pages={80--91},
  year={1959},
  publisher={Informs}
}

@article{reimers2019sentence,
  title={Sentence-bert: Sentence embeddings using siamese bert-networks},
  author={Reimers, Nils and Gurevych, Iryna},
  journal={arXiv preprint arXiv:1908.10084},
  year={2019}
}

@article{Funsearch,
  title={Mathematical discoveries from program search with large language models},
  author={Romera-Paredes, Bernardino and Barekatain, Mohammadamin and Novikov, Alexander and Balog, Matej and Kumar, M Pawan and Dupont, Emilien and Ruiz, Francisco JR and Ellenberg, Jordan S and Wang, Pengming and Fawzi, Omar and others},
  journal={Nature},
  volume={625},
  number={7995},
  pages={468--475},
  year={2024},
  publisher={Nature Publishing Group UK London}
}

@article{EoH,
  title={Evolution of heuristics: Towards efficient automatic algorithm design using large language model},
  author={Liu, Fei and Tong, Xialiang and Yuan, Mingxuan and Lin, Xi and Luo, Fu and Wang, Zhenkun and Lu, Zhichao and Zhang, Qingfu},
  journal={arXiv preprint arXiv:2401.02051},
  year={2024}
}
 
\end{document}